\documentclass[conference]{IEEEtran}
\pdfoutput=1
\IEEEoverridecommandlockouts
\usepackage{cite}
\usepackage{amsmath,amssymb,amsfonts}
\usepackage{algorithmic}
\usepackage{graphicx}
\usepackage{textcomp}
\usepackage{xcolor}
\def\BibTeX{{\rm B\kern-.05em{\sc i\kern-.025em b}\kern-.08em
    T\kern-.1667em\lower.7ex\hbox{E}\kern-.125emX}}
    
\usepackage{hyperref}
\begin{document}

\title{An AIoT-enabled Autonomous Dementia Monitoring System\\
\thanks{The code of this work is publicly available here: \href{https://github.com/wxy12151/IoT-system-for-detecting-the-signs-of-deterioration-in-the-person-with-a-chronic}{Github Link}}
}

\author{\IEEEauthorblockN {Jinyang Li and Xingyu Wu}
\IEEEauthorblockA{Dept. of Electrical and Electronic Engineering \\
University College London\\
London, United Kingdom \\
Email: \href{mailto:jinyang.li.21@ucl.ac.uk}{jinyang.li.21@ucl.ac.uk} and  \href{mailto:x.wu.20@ucl.ac.uk}{x.wu.20@ucl.ac.uk}}
}

\maketitle

\begin{abstract}
An autonomous Artificial Internet of Things (AIoT) system for elderly dementia patients' monitoring in a smart home is presented. The system mainly implements two functions based on the activity inference of the sensor data, which are real-time abnormal activity monitoring and trend prediction of disease related activities.
Specifically, CASAS\cite{cook2012casas} dataset is employed to train a Random Forest (RF) model for activity inference. Then, another RF model trained by the output data of activity inference is used for abnormal activity monitoring.
Particularly, RF is chosen for these tasks because of its balanced trade-offs between accuracy, time efficiency, flexibility, and interpretability. 
Moreover, Long Short-Term Memory (LSTM) is utilised to forecast the disease-related activity trend of a patient.
Consequently, the accuracy of two RF classifiers designed for activity inference and abnormal activity detection is greater than 99\% and 94\%, respectively. Furthermore, using the duration of a patient's sleep as an example, the LSTM model achieves accurate and evident future trends prediction.

\end{abstract}


\section{Introduction}
Dementia is employed as a chronic health condition to be monitored in our AIoT system. Generally, it is a prevalent occurrence in older folks \cite{livingston2020dementia}\cite{cao2020prevalence}.
There is a lot of concern about the well-being of the elderly in many regions of the world nowadays. Therefore, caregivers and healthcare centers can benefit from remote monitoring of elderly folks' health and well-being.
\cite{johannessen2018or} conducted interviews with dementia patients to investigate their behaviors during flare-ups and exacerbation.
According to the interview, memory loss, abnormal noise during sleep, insomnia and night wandering are adopted as symptoms to be monitored in this work. And these symptoms could be embodied in the daily activities such as sleep, work, personal hygiene, etc \cite{khan2018detecting}\cite{thorpe2019development}\cite{arifoglu2019detection}.


To monitor dementia patients' health status intelligently from their daily activities, the AIoT-enabled autonomous system is deployed, achieving the ultimate goal of real-time abnormal activity monitoring and trend prediction of disease-related activities. Specifically, two RF models are employed for both tasks with high interpretability. 

\begin{figure}[htb]
\centerline{\includegraphics[width=0.75\columnwidth]{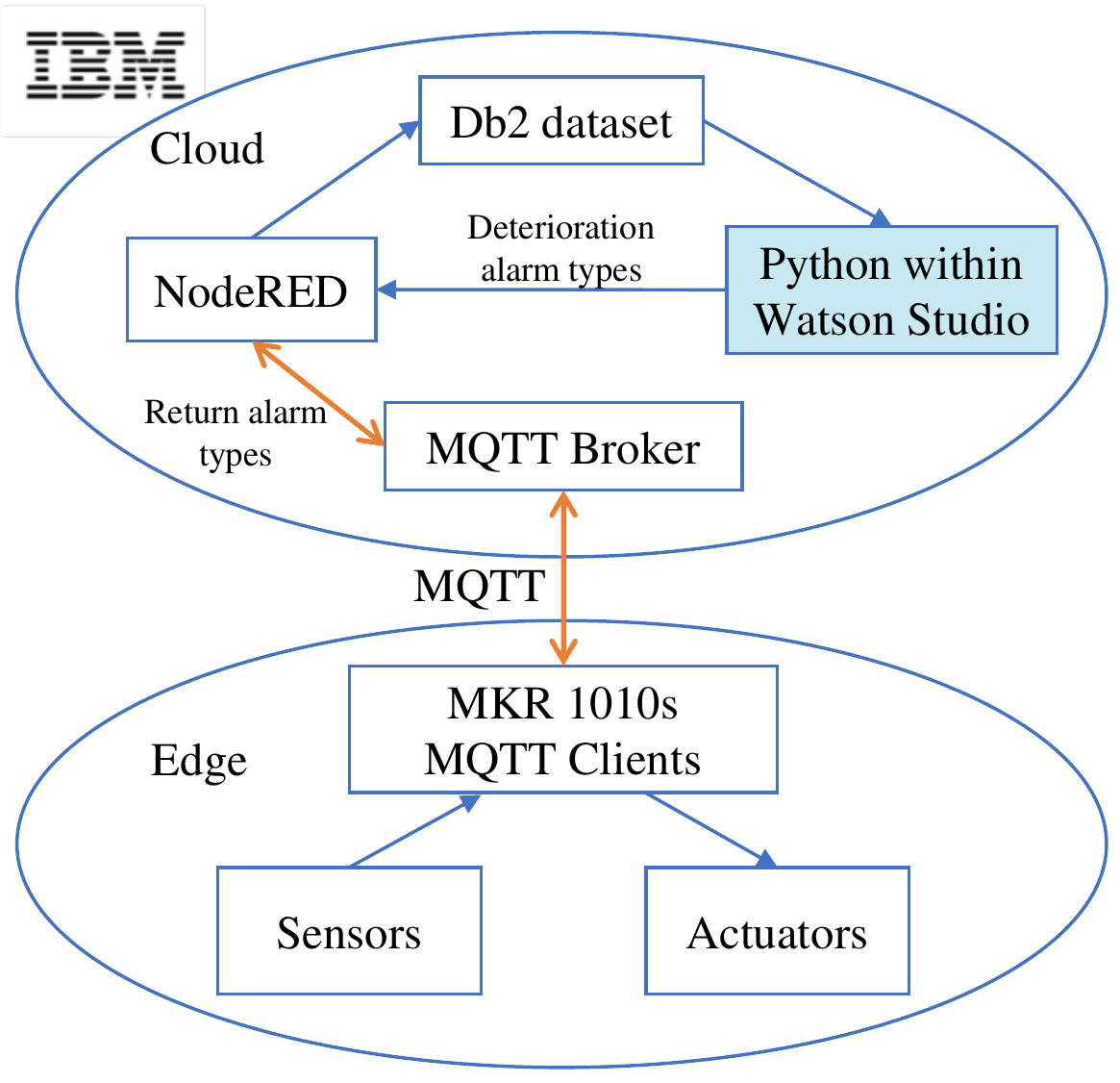}}
\caption{The scheme of the AIoT System.}
\label{fig:iot stack}
\end{figure}

To complete this end-to-end AIoT system, as seen in Figure \ref{fig:iot stack}, the edge system, connectivity and cloud must all be considered. The edge system needs to gather patient activity data via sensors and send various types of reminders to roommates or medical teams via actuators. Meanwhile, some connection protocols (e.g., MQTT) are employed to ensure stable connections. Particularly, NodeRED on IBM is utilized to monitor the disease's progression and provide an interface, and this work will focus on the "Python within Watson Studio" part on the cloud, which involves the mechanisms to detect and anticipate the disease-related activity.

\section{Methodology}
\subsection{Scheme Overview}
\begin{figure}[htb]
\centerline{\includegraphics[width=0.75\columnwidth]{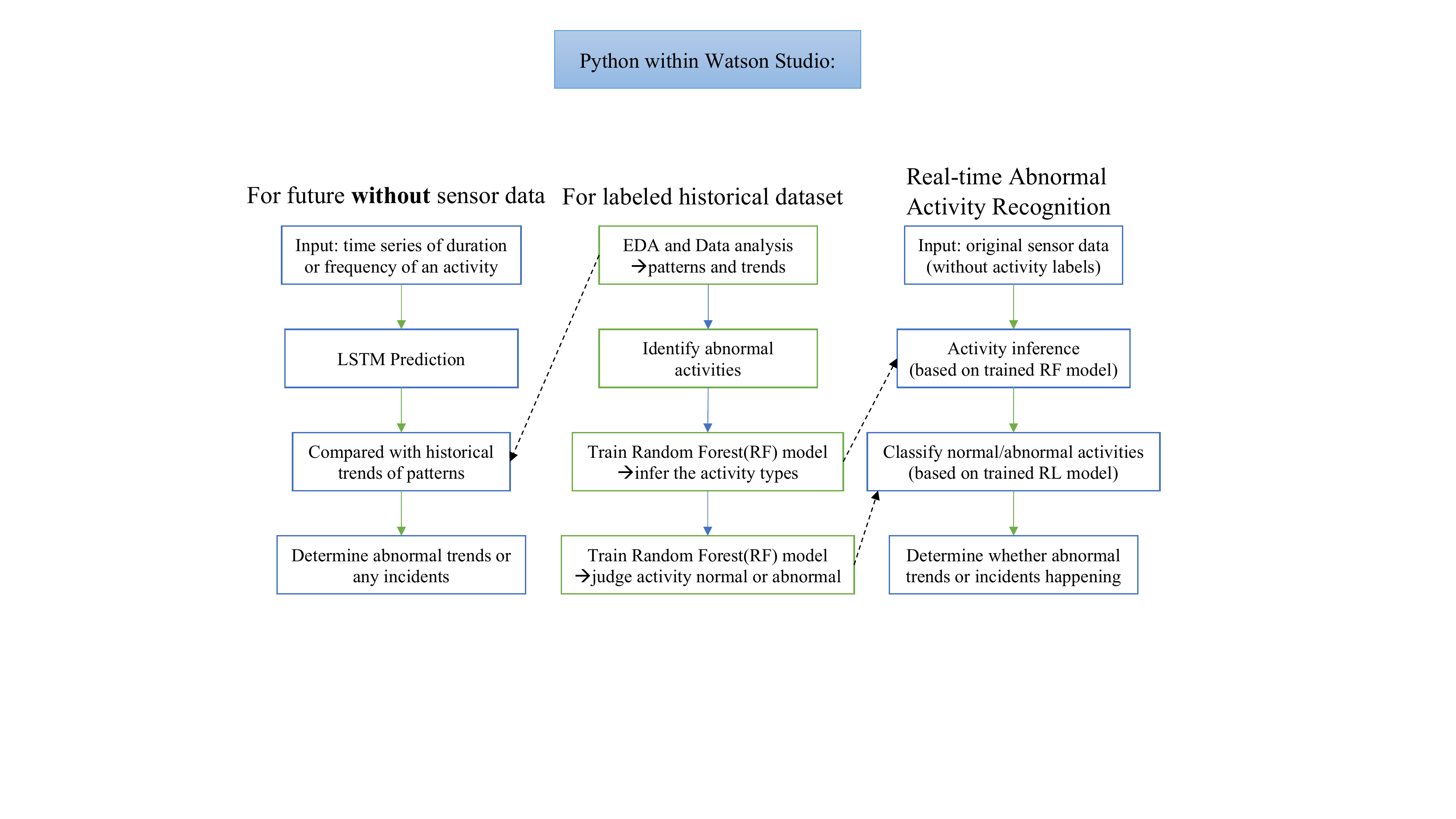}}
\caption{Scheme overview of the AIoT-enabled Dementia Monitoring System.}
\label{fig:phaseIISchema}
\end{figure}

As Figure \ref{fig:phaseIISchema} shows, the main functions of the system are real-time abnormal activity inference, and activity trends prediction for the dementia patient, which needs the patient's activity data as the input. The activity data is produced by activity inference on the raw data uploaded from the sensors located in the patient's living environment. 

\begin{figure}[htb]
        \centerline{\includegraphics[width=\columnwidth]{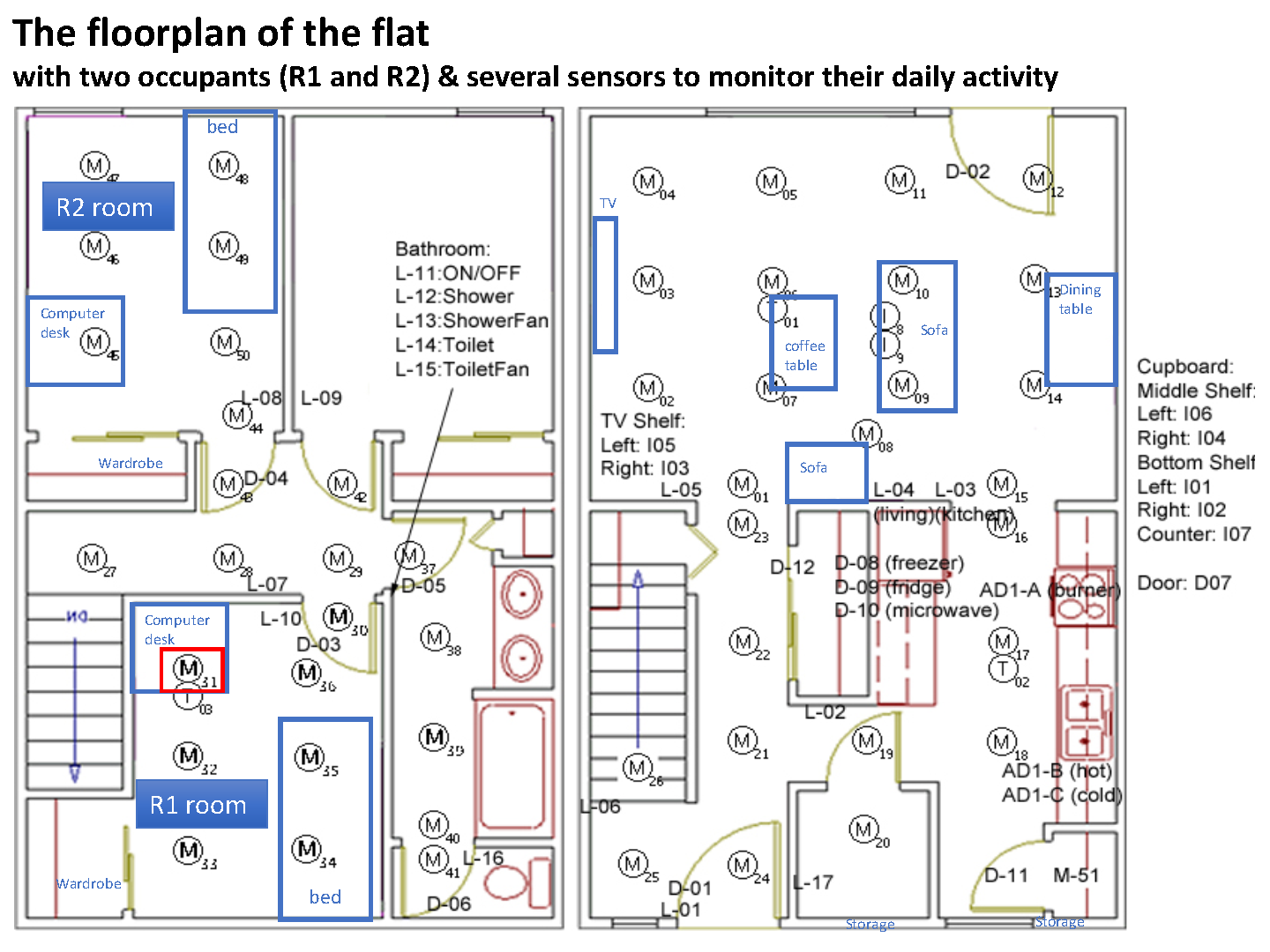}}
        \caption{Floor plan of the smart home related to CASAS dataset.}
        \label{fig:floorplan}
        \end{figure}

To assess our approach, the CASAS dataset, collected in a smart home environment \cite{cook2012casas}, is utilized in this project, which contains the recordings uploaded from several sensors that were deployed in a flat with two residents, for a period of about $3$ months. 
        Specifically, the locations of the sensors are shown in Figure \ref{fig:floorplan}. The sensors deployed can be categorized as motion sensors, kitchen item sensors, door sensors, light sensors, AD1-A for burner sensors, AD1-B for hot water sensors, and AD1-C for cold water sensors. They monitor the daily activities of the two residents, such as R1's bed to toilet, R1's personal hygiene, meal preparation, etc.

\subsection{Sensor-based Activity Inference}
To infer the patient's real-time activities from the corresponding streaming sensor data, an ML-based activity inference approach was designed containing the steps of \textbf{sensor data windowing}, \textbf{weighting}, \textbf{data transformation} and \textbf{activity classification}.

\subsubsection{Sensor Data Windowing}
Consideration of preceding sensor events windowing is a straightforward method for providing context to determine what activity a single sensor event belongs to. 
The windows were set up with a sensor-based approach where data is partitioned into equal-number-of-sensor-events windows \cite{gu2009epsicar}.

\subsubsection{Sensor Weighting}
The goal of sensor data windowing is to provide effective context information for the last sensor event in the window.  However, there may exist disruptive sensor events that are labelled with other activities than that of the last sensor event in the window. To deal with this problem, a sensor weighting matrix is built with reference to the sensor dependency weighting scheme proposed by Yala et al. \cite{yala2017towards} based on a mutual information measure between the sensors. Specifically, the mutual information between two sensors $S_{i}$ and $S_{j}$ is calculated by their frequency of occurrence in a sequence of sensor events representing a single activity as the following equation:

        \begin{small}
        \begin{equation}
        M I(S_{i}, S_{j})=\frac{1}{|Q|} \sum_{k=1}^{|Q|} \partial(S_{i}, S_{j})
        \end{equation}
        \end{small}
        
where $\partial(S_{i}, S_{j}) = 1$ if sensors $S_{i}$ and $S_{j}$ are both activated during the sequence of the sensor events in an activity. And $Q$ represents the total number of dataset activities.

\subsubsection{Data Transformation}
On the basis of sensor data windowing and sensor weighting, effective information from multiple sensor events in each window containing the weighted occurrence times of each sensor, the end time and duration of the window, and the last two sensor IDs will be derived and the window will be transformed into a single aggregated data item. 

\subsubsection{Activity Inference using Random Forest}
Machine Learning (ML) could generate a learning-based intelligent model that learns from the labelled training dataset output by the previous steps to determine the activity type, which generally has better overall performance 
compared with rule-based methods \cite{theekakul2011rule}. Among various learning-based ML models, we chose Random Forest (RF) \cite{biau2016random} because of its \textbf{high accuracy}, \textbf{flexibility} (suitable for different data distributions), \textbf{interpretability}, and \textbf{low running time complexitiy}. 
RF is made up of a couple of decision trees (DTs) \cite{myles2004introduction} as the basic predictors. DT is a heuristic ML technique that recursively and greedily assesses the features of the input data. Specifically, a decision tree splits its branch by testing for one of the input data’s properties. And every step conducts testing based on the divisions established in the preceding steps. Within a tree, the divide-and-conquer approach divides the entire input dataset into several portions that must be identified from root to leaves. Each path from the root to a leaf node corresponds to classifying a portion of the dataset. The core algorithm DT uses is Gini Index. Assume the proportion of samples that belong to the k-th class in a dataset is $p_{k} (k=1,2,...,m)$. Then, we can measure the purity of dataset D with the Gini value as:
\begin{equation}
\label{equ_gini}
\begin{aligned}
\operatorname{Gini}(D) &=\sum_{k=1}^{m} \sum_{k^{\prime} \neq k} p_{k} p_{k^{\prime}} 
\end{aligned}
\end{equation}

\begin{figure}[htb]
\centerline{\includegraphics[width=0.7\columnwidth]{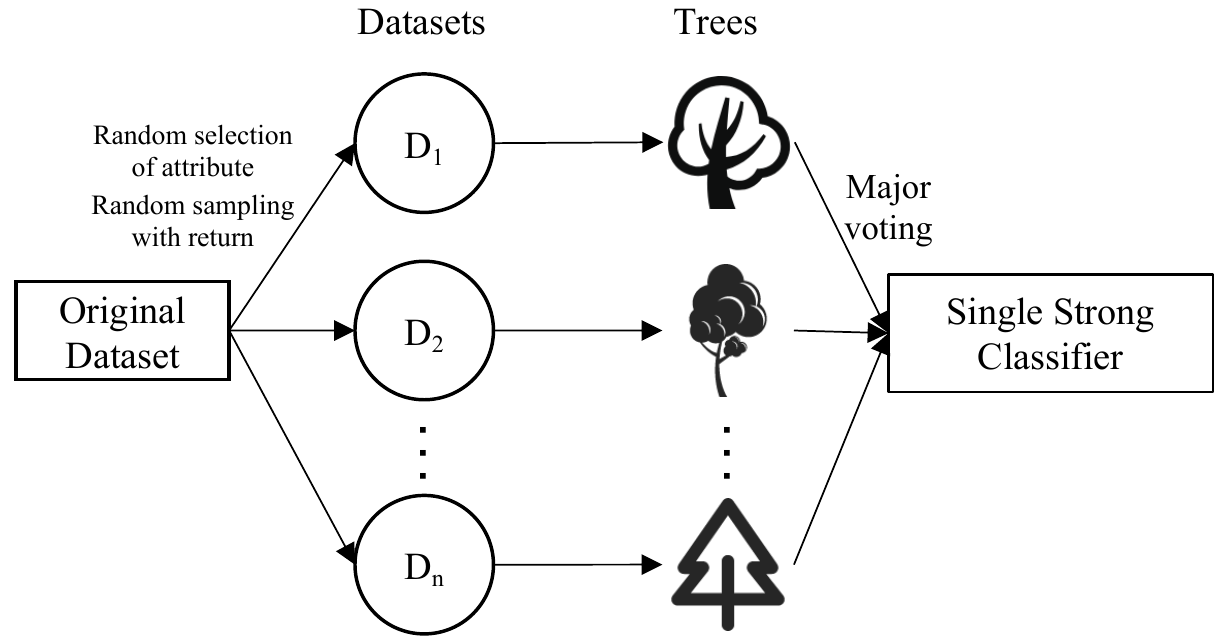}}
\caption{The architecture of Random Forest.}
\label{fig:RFStructure}
\end{figure}

RF, as illustrated in Figure \ref{fig:RFStructure}, is a decision tree ensemble model proposed to solve the problem of sticking in the local optimal of a single tree. It trains a number of decision trees using the method of randomly selecting subsets of the attribute set based on the bagging strategy. In RF, bagging is the process of randomly selecting multiple different subsets of the training set and fitting several decision trees individually. The results are then obtained by major voting all trees. Generally, RF can increase classification accuracy and decrease over-fitting compared with a single decision tree.

\subsection{Real-time Abnormal Activity Recognition}
As Figure \ref{fig:phaseIISchema} shows, the patient's activity data can be acquired with the previous processes. Moreover, a set of related statistical data concerning the start and end time, time span, and happening sequence of activity on a daily basis can be acquired. Taking this extracted activity statistical data as input, a classifier that determines if such a data item is abnormal can be built as a necessary part of functionalities for the goal of detecting the dementia status. 

To train an ML model in a supervised way, the activity statistical data items are supposed to label with ``normal'' or ``abnormal''. Z-score, a popular statistical method for abnormal data detection, is mainly utilized here to determine the abnormal quantitative attributes of frequency and time span. Furthermore, the abnormal start time and end time are determined by setting an abnormal time range for each activity according to Exploratory Data Analysis (EDA).

After labelling the activity dataset, an RF classifier for abnormal activity is developed and evaluated based on its balanced trade-offs.

\subsection{Activity Trends Prediction} 
As an important functionality that reminds the risk of deterioration without future sensor data input, a time-series prediction model needs to be built based on historical trends of activity data. 

Among the various AI models for time-series prediction, Long Short-Term Memory (LSTM) \cite{hochreiter1997long} has been evaluated as being able to deliver a general high performance in different applications \cite{liu2019analysis}\cite{rahhal2020iot}, which is employed in this task.

\begin{figure}[htb]
\centerline{\includegraphics[width=0.7\columnwidth]{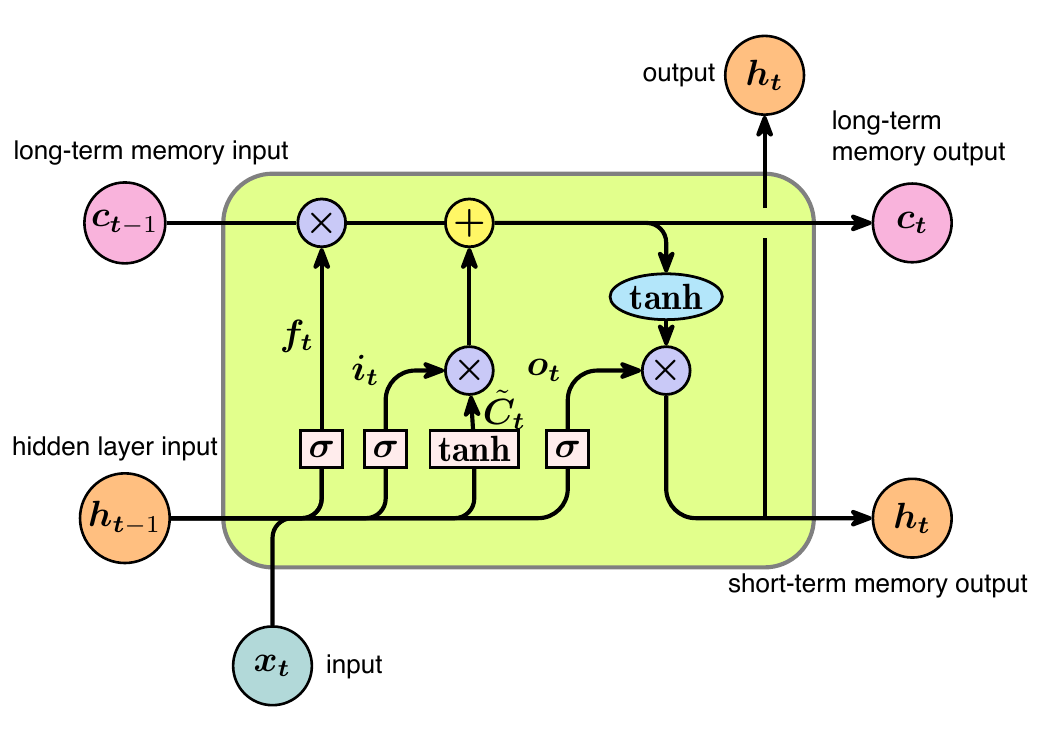}}
\caption{The architecture of LSTM.}
\label{fig:LSTMStructure}
\end{figure}

LSTM uses gate control to keep the useful past states and abandon the useless ones. As Figure \ref{fig:LSTMStructure} shows, in an LSTM network, each unit consists three gates: input gate $i_{t}$, forget gate $f_{t}$, and output gate $o_{t}$. The following is how the gates work: To determine this memory's state $c_{t}$, an LSTM unit takes the previous memory state $c_{t-1}$ and performs element-wise multiplication with the forget gate $f$. The prior memory state is fully forgotten if the forget gate value is 0. Otherwise, the current unit receives the entire previous memory state. The calculation of the states in a unit is shown in Equation \ref{equ:LSTMCalculate}.

\begin{equation}
\label{equ:LSTMCalculate}
\begin{aligned}
f_{t} &=\sigma\left(W_{f h} h_{t-1}+W_{f x} x_{t}+b_{f}\right) \\
i_{t} &=\sigma\left(W_{i h} h_{t-1}+W_{i x} x_{t}+b_{i}\right) \\
\tilde{C}_{t} &=\tanh \left(W_{\tilde{c} h} h_{t-1}+W_{\tilde{c} x} x_{t}+b_{\tilde{c}}\right) \\
c_{t} &=f_{t} \cdot c_{t-1}+i_{t} \cdot \tilde{C}_{t} \\
o_{t} &=\sigma\left(W_{o h} h_{t-1}+W_{o x} x_{t}+b_{o}\right) \\
h_{t} &=o_{t} \cdot \tanh \left(c_{t}\right)
\end{aligned}
\end{equation}



\section{Results and Discussion}
\subsection{Activity Recognition Results}
The results obtained from the RF model trained on the streaming sensor data indicate a high performance, with a testing accuracy of 99.13\%, with an 8:2 ratio of training to the testing dataset. Besides, as seen in Table \ref{tab:ARandAbnormalResults}, other metrics of precision, recall, f1-score, and AUC all maintain a weighted average value higher than 99\%. They show the effectiveness of RF in activity recognition.
\subsection{Abnormal Activity Detection Results}
\begin{figure}[htb]
\centerline{\includegraphics[width=0.85\columnwidth]{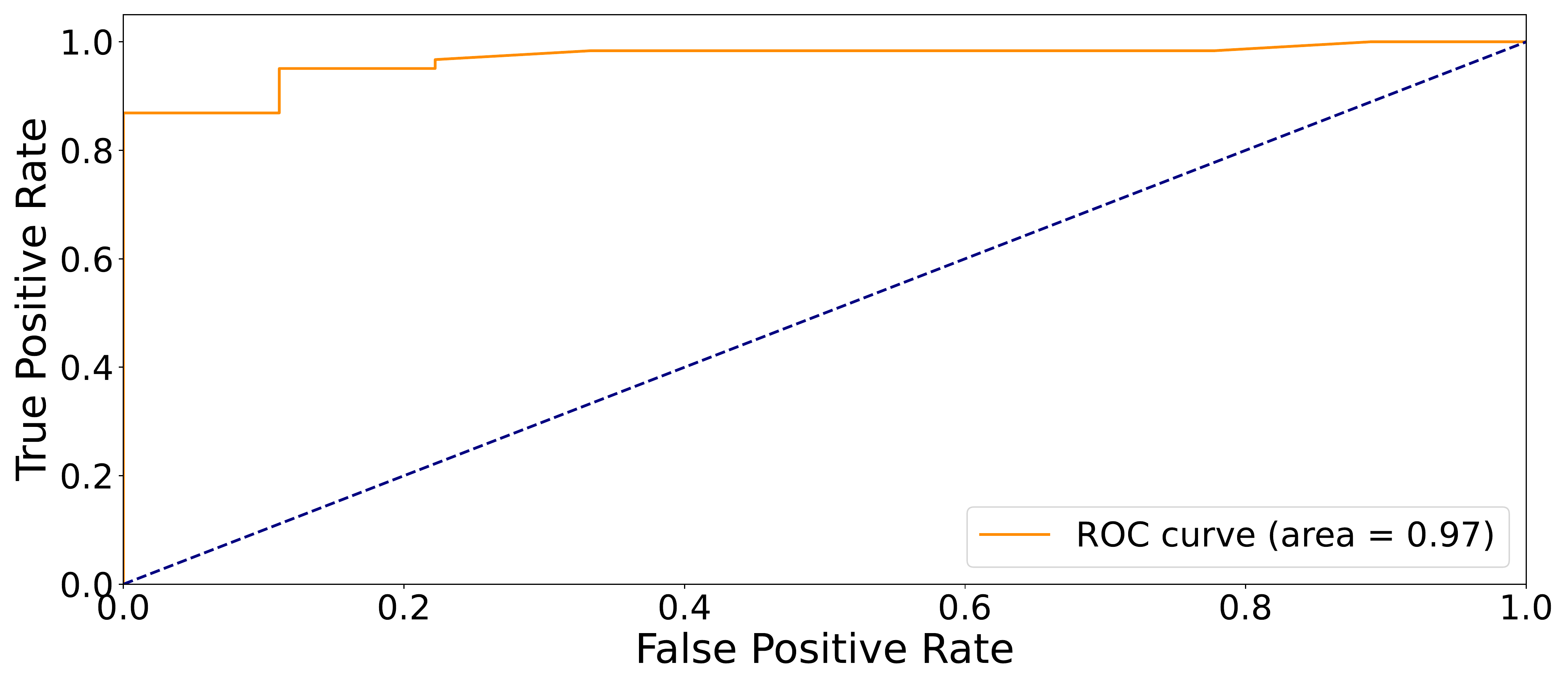}}
\caption{ROC curve of RF model for abnormal activity identification.}
\label{fig:ROCAbnormalIdentification}
\end{figure}

The results obtained from the RF model trained on the activity data indicate a relatively high testing accuracy of 94.29\%. 
Besides, as Table \ref{tab:ARandAbnormalResults} shows, other metrics of precision, recall, and f1-score maintain a similar weighted average value higher than 94\%, and an AUC value of 0.97, as shown in Figure \ref{fig:ROCAbnormalIdentification}. 

\linespread{1.5}
\begin{table}[htb]
\centering
\caption{Metrics performances of RF models on activity recognition and abnormal activity identification.}
\label{tab:ARandAbnormalResults}
\resizebox{\columnwidth}{!}{%
\begin{tabular}{c|cccc}
\hline
 &
  \textbf{precision} &
  \textbf{recall} &
  \textbf{f1-score} &
  \textbf{AUC} \\ \hline
\textbf{Activity Recognition} &
  0.99128855 &
  0.99129964 &
  0.99126821 &
  1.00 \\ \hline
\textbf{\begin{tabular}[c]{@{}c@{}}Abnormal   Detection\end{tabular}} &
  \multicolumn{1}{l}{0.94285714} &
  \multicolumn{1}{l}{0.94285714} &
  \multicolumn{1}{l}{0.94285714} &
  0.97 \\ \hline
\end{tabular}%
}
\end{table}
\linespread{1}

\subsection{Activity Trends Prediction Results}
\begin{figure}[htb]
\centerline{\includegraphics[width=1\columnwidth]{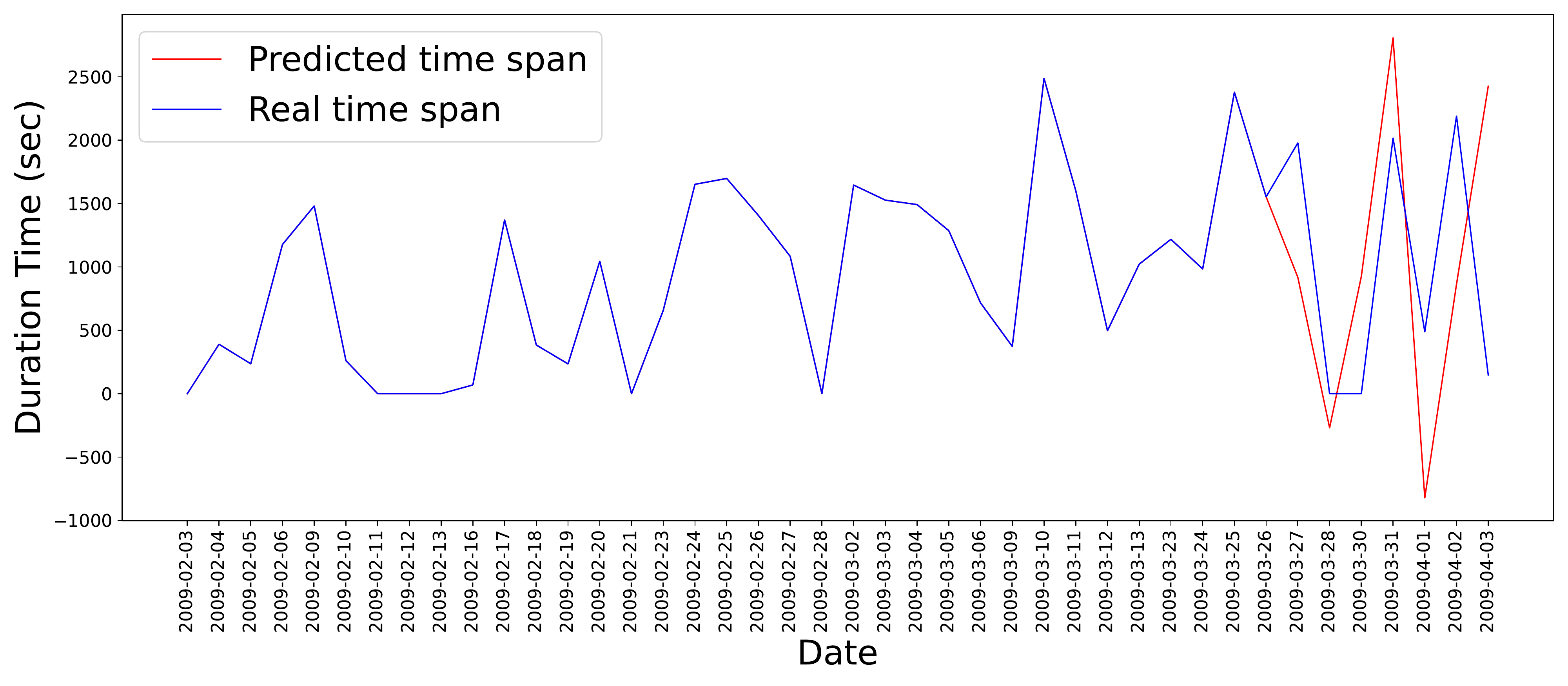}}
\caption{LSTM prediction results on R1's future personal hygiene duration.}
\label{fig:LSTMResults}
\end{figure}

An LSTM model is trained for the prediction of the patient's activities' future frequencies and time duration on a weekly basis. One example is shown in Figure \ref{fig:LSTMResults}. The model takes the historical time duration data of 21 days as input and outputs the prediction for the future 7 days. The prediction results on the testing data shown by the red line have proved the model's ability to predict future trends and generate appropriate risk alerting of potential abnormal activities for the caregivers. 
Intuitively, the progression of the patient's condition can be determined by comparing the expected and actual curves.

\subsection{RF Interpretability Evaluation}
\begin{figure}[htb]
\centerline{\includegraphics[width=0.8\columnwidth]{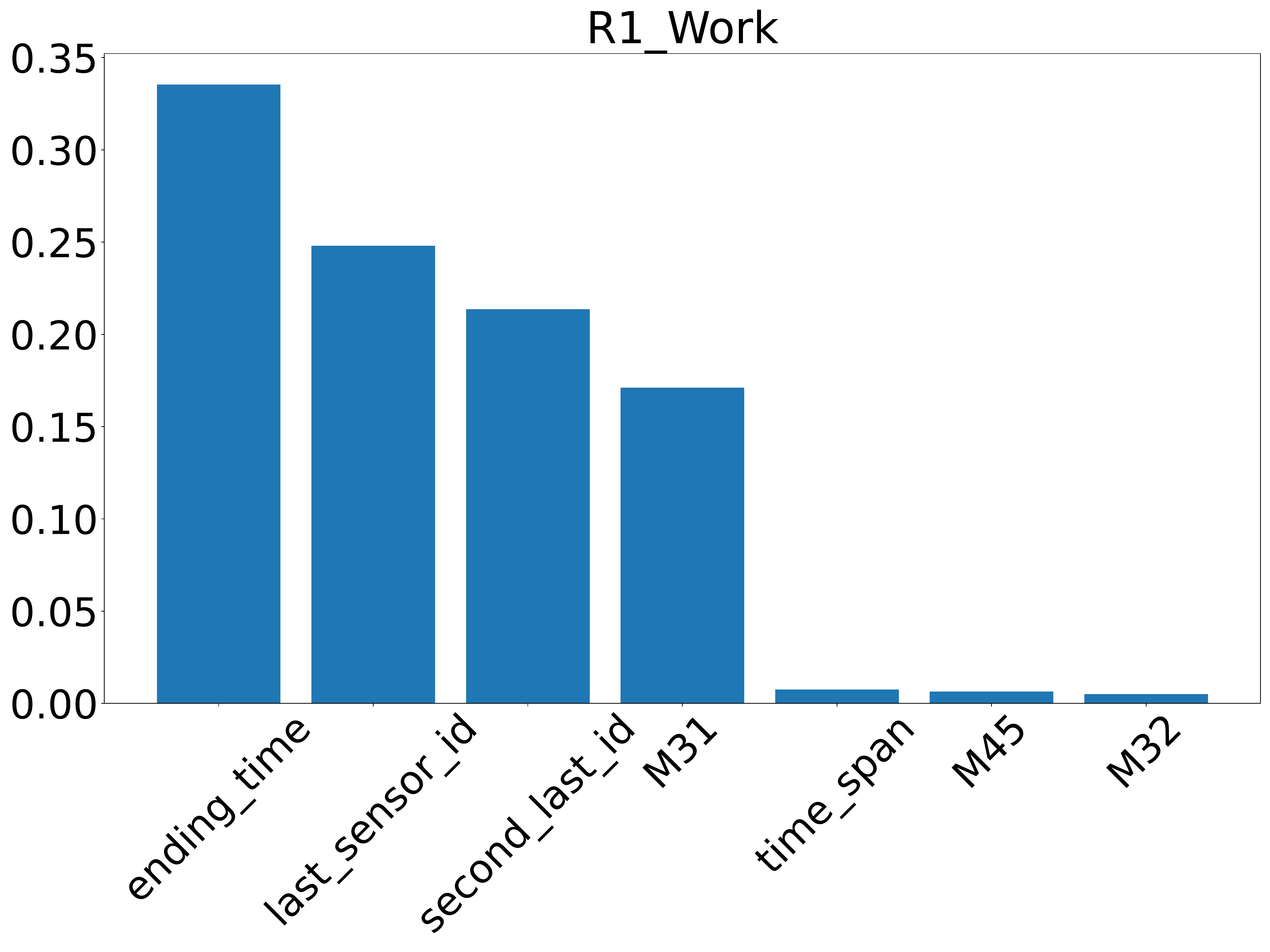}}
\caption{First 7 important features of the activity R1\_work. }
\label{fig:R1WorkImportance}
\end{figure}

Different from most deep-learning-based AI models which are black boxes, RF can provide higher interpretability to earn more user trust and achieve high accuracy and time efficiency at the same time. One of the approaches for RF enhancing the interpretability is to quantify the importance of each data feature for the task. Specifically, for activity inference, RF can quantify the feature importance corresponding to the different patient activity type. For example, Figure \ref{fig:R1WorkImportance} shows the first $7$ important features for determining the input data item as ``R1\_work'', in which the ending time of the window, last two sensors' IDs, and the triggering of the motion sensor M31 are essential signs of R1\_work. 
Particularly, it is visible from Figure \ref{fig:floorplan}'s floor plan that the motion sensor M31 is positioned on the computer desk of the patient's room, which can be frequently activated when the patient is working.

\begin{figure}[htb]
\centerline{\includegraphics[width=0.8\columnwidth]{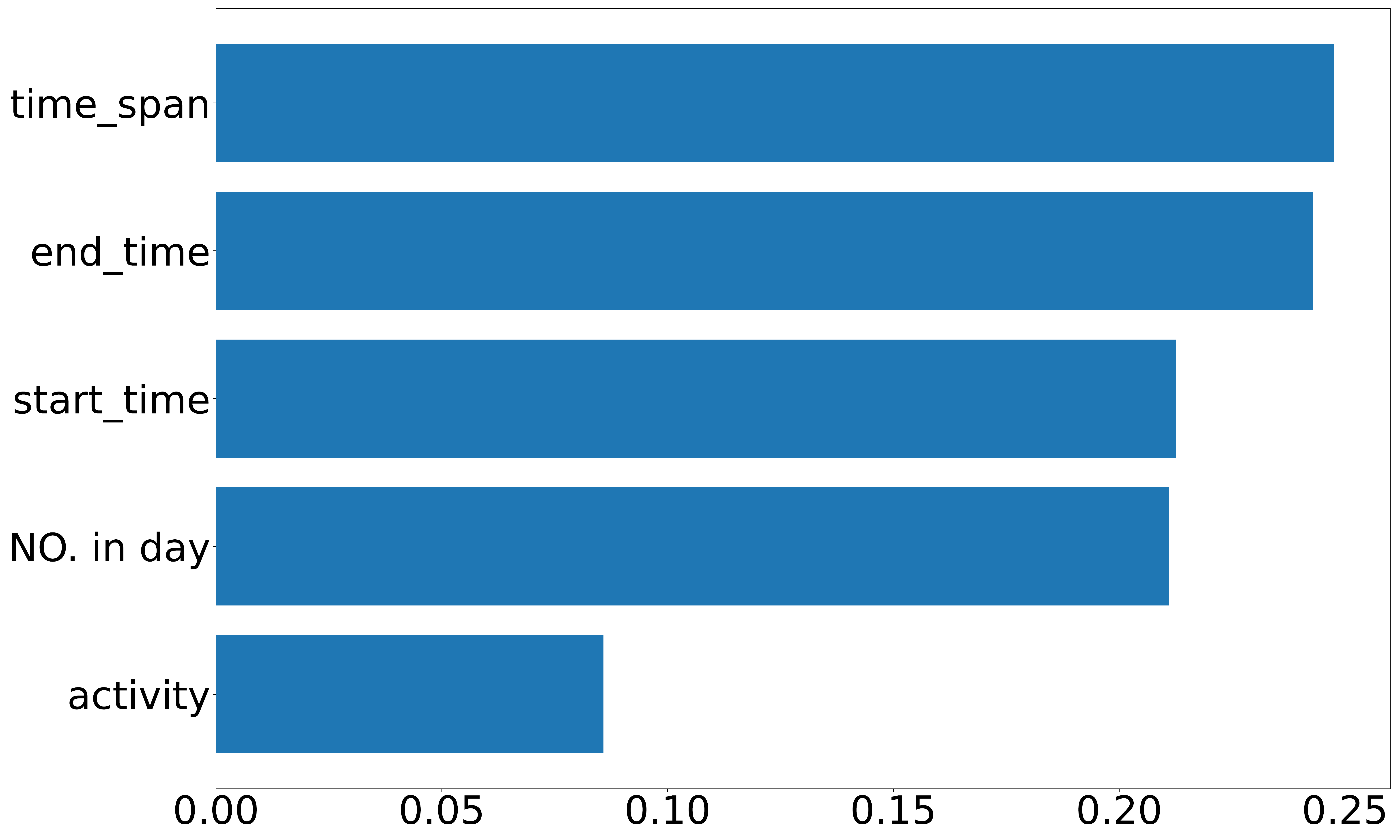}}
\caption{Feature importance in abnormal activity detection.}
\label{fig:AbnormalFeatureImporatnce}
\end{figure}

Moreover, for abnormal activity detection, Figure \ref{fig:AbnormalFeatureImporatnce} shows the feature importance rank, with the time span of an activity being the most important feature, followed by the activity ending time, starting time, happening sequence in the day, and the activity type.

\section{Conclusion}
This article presents an autonomous AIoT-enabled
system for elderly dementia patients’ monitoring in a smart home. Specifically, sensor-based activity inference, real-time abnormal recognition, activity trends prediction are mainly proposed in this work.
Listed below are the system's benefits:\\
(1) The system can be easily deployed without manual participation after training the related AI models. \\ 
(2) Using the context of previous sensor events, the system can automatically identify the type of patient activity to which a newly arriving sensor event belongs. \\
(3) The system can automatically determine whether a patient's present activity is normal based on the activity's duration and daily frequency, etc., thereby achieving real-time healthcare monitoring.\\
(4) 
The system can forecast automatically the relevant characteristics of dementia-related activities, such as sleeping time, and alert the healthcare provider in advance.

Consequently, the system achieves an accuracy of 99.13\% in activity recognition and 94.29\% in the detection of abnormal activity on the CASAS \cite{cook2012casas} dataset.

\clearpage

\bibliographystyle{IEEEtran}


\clearpage


\end{document}